\documentclass[10pt,twocolumn,letterpaper]{article}

\usepackage[pagenumbers]{cvpr} 
\usepackage{booktabs}
\usepackage{array} 
\usepackage{multirow} 
\usepackage{xr} 

\definecolor{cvprblue}{rgb}{0.21,0.49,0.74}
\newcommand{\PAR}[1]{\noindent{\bf #1}}

\usepackage[pagebackref,breaklinks,colorlinks,allcolors=cvprblue]{hyperref}

\newcommand*{\fp}{focus position\@\xspace}

\def\paperID{633} 
\def\confName{CVPR}
\def\confYear{2025}

\title{One-Step Event-Driven High-Speed Autofocus}

\author{
    Yuhan Bao \hspace{3em} Shaohua Gao \hspace{3em} Wenyong Li \hspace{3em} Kaiwei Wang*\\
    State Key Laboratory of Extreme Photonics and Instrumentation, Zhejiang University, China\\
    {\tt\small \{yhbao, gaoshaohua, liwenyong2023, wangkaiwei\}@zju.edu.cn}
}

\begin{document}
\maketitle
\begin{abstract}
High-speed autofocus in extreme scenes remains a significant challenge. Traditional methods rely on repeated sampling around the focus position, resulting in ``focus hunting''. Event-driven methods have advanced focusing speed and improved performance in low-light conditions; however, current approaches still require at least one lengthy round of ``focus hunting'', involving the collection of a complete focus stack. We introduce the Event Laplacian Product (ELP) focus detection function, which combines event data with grayscale Laplacian information, redefining focus search as a detection task. This innovation enables the first one-step event-driven autofocus, cutting focusing time by up to two-thirds and reducing focusing error by 24 times on the DAVIS346 dataset and 22 times on the EVK4 dataset. Additionally, we present an autofocus pipeline tailored for event-only cameras, achieving accurate results across a range of challenging motion and lighting conditions. All datasets and code will be made publicly available.

\end{abstract}    
\section{Introduction}
\label{sec:intro}
Focus is a prerequisite for most visual tasks. An ideal autofocus (AF) system guides the motor directly toward the correct focus position from any starting point, and stop immediately upon arrival, ensuring precision and efficiency, which can be considered as ``one-step AF''.

Conventional AF algorithms predict focus by evaluating the image contrast~\cite{zhang2018autofocus} (or sharpness) at different points, which often requires repeated sampling around the focus position, leading to ``focus hunting''. At the same time, the contrast-based AF methods are susceptible to low frame rates, motion blur, and so on. To enhance speed and accuracy, Phase Detection AutoFocus (PDAF) with dual-pixel sensors is commonly employed. These sensors measure phase differences to calculate focus position. However, PDAF is confronted with several challenges: the complexity of pixel design limits the number of dual pixels, reduced light intake impacts performance in low-light conditions, and it struggles in scenarios with significant defocus.

Recent research has shown the potential of event cameras for high-speed AF applications. Event cameras detect brightness changes at individual pixels with extreme temporal resolution, down to 1 $\mu$s, making them ideal for dynamic scenarios. However, the unique asynchronous characteristics of event cameras set event-driven AF methods apart from traditional frame-based approaches.
The use of event rate (ER) as a focus evaluation function was first proposed in \cite{lin2022autofocus}, where the \fp is identified by locating the position with the highest ER in the focus stack. Building on this, the Event Golden Search (EGS) algorithm was introduced to improve the speed of focus search. 
Additionally, another study~\cite{bao2023improving} noted that brightness changes exhibit symmetry around the \fp in the focus stack. Building on this insight, the Polarity Based Autofocus (PBF) algorithm was developed, effectively leveraging this symmetry to achieve rapid and precise focusing across various lighting conditions, including strobe lighting.
However, current event-driven AF algorithms require analyzing a complete focus stack, including information both before and after the \fp, before the \fp can be determined. While these methods eliminate repeated ``focus hunting'', they still require a single round of it, which involves capturing the entire focus stack, searching for the \fp, and moving there.
\begin{figure}[t]
  \centering
   \includegraphics[width=0.99\linewidth]{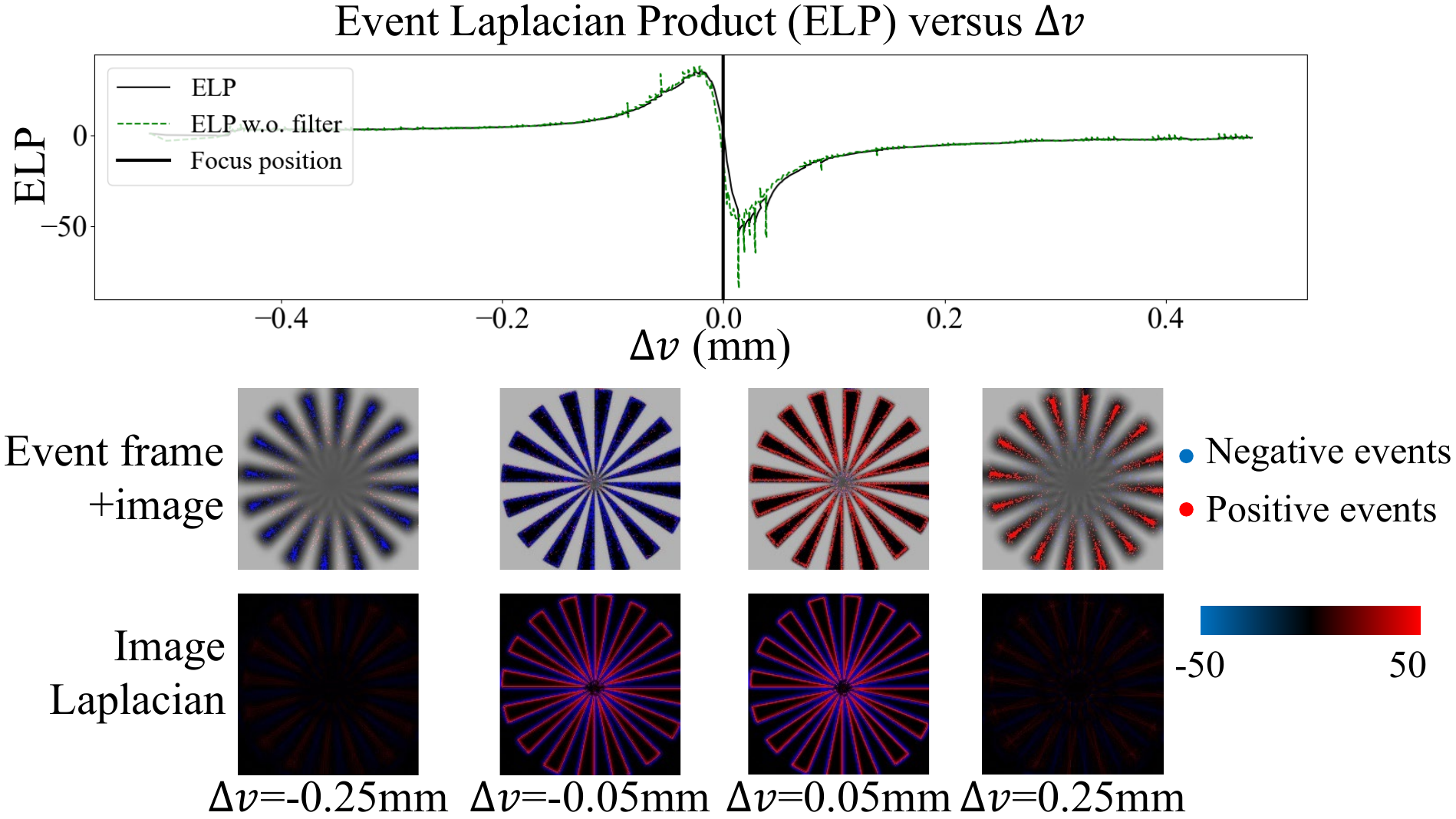}
   \caption{Visualization of ELP, event frames, images and Laplacian at different $\Delta v$.}
   \label{fig:ELP}
\end{figure}

For the first time, we integrate both events and grayscale information to achieve one-step event-driven AF. We theoretically derive a fundamental connection between the spatial second-order derivative and the temporal first-order derivative of the image during the focusing process. Building on this, we introduce a focus \textbf{detection} function called the ``Event Laplacian Product'' (ELP), which shows a distinct ``sign mutation'' at the \fp, as shown in~\cref{fig:ELP}.
The proposed ELP method determines the focus position in real time by detecting the ``sign mutation'' of the ELP value from positive to negative, eliminating the need for iterative searches required by traditional methods based on the ``peaked'' focus evaluation function. Additionally, ELP predicts the direction of focus adjustment based on the sign of its value, effectively serving as a focus \textbf{prediction} function.
Moreover, by integrating ELP with the latest event-based temporal mapping photography~\cite{bao2024temporal}, we achieve one-step high-speed AF using event only.

To the best of our knowledge, we are the first to implement event-driven one-step AF, achieving accurate focusing with less than one depth of focus across various brightness conditions and motion states in the synthetic, DAVIS346, and Prophesee EVK4 datasets. Compared to existing state-of-the-art event-driven AF method, ELP reduces focusing error by \textbf{24 times} on the DAVIS346 dataset and by \textbf{22 times} on the EVK4 dataset. Additionally, the focusing time is reduced by \textbf{two-thirds}.
Compared to the current one-step AF method PDAF, our approach is not limited by the number of dual-pixels, remains robust in low-light conditions, and performs reliably even under significant defocus.

\section{Related Work}
\label{sec:related}
\subsection{Conventional camera AF}
Currently, the widely-used AF methods for conventional cameras are contrast-based AF and PDAF, or a mixture of both.
The contrast-based AF algorithm consists of two key components: the focus evaluation function and the search algorithm. The focus evaluation function typically includes: (1) First-order gradient methods, such as the Prewitt operator~\cite{lofroth2018auto} and Sobel operator~\cite{mo2012auto}, (2) Second-order gradient methods, like the Laplacian operator~\cite{jia2022autofocus}, and (3) Histogram-based methods~\cite{guo2018fast}, among others. Search algorithms commonly used include hill climbing~\cite{he2003modified} and Fibonacci search~\cite{krotkov1988focusing}. More recently, deep learning-based AF approaches have shown promising results~\cite{pinkard2019deep}, offering some ability to predict the focus position. Yang~\cite{yang2022deep} introduced Differential Focus Volume into the depth from focus field, where the concept of differentiation also serves as an inspiration for AF tasks.
However, a significant issue with contrast-based AF is the ambiguity between pre-focus and post-focus, which has been addressed by the introduction of PDAF~\cite{sliwinski2013simple}. By analyzing phase differences between the two elements of dual-pixel sensors, PDAF can determine both the direction and the distance needed for focusing, enabling one-step AF. Nevertheless, the complexity of dual-pixel design and its impact on image quality limit the number of dual-pixel sensors in digital cameras. In conventional camera AF, the frame rate for acquiring raw image data has long been a limiting factor for AF speed. Furthermore, the problem of ``focus hunting'' in complex scenes—such as low light, motion, and significant defocus—remains a persistent challenge.

\subsection{Event-driven AF}
The high temporal resolution, dynamic range, and low data redundancy of event cameras offer new possibilities for AF and related applications~\cite{teng2024hybrid,ralph2024active}.
The use of ER as a focus evaluation function for event cameras was first introduced in \cite{lin2022autofocus}, along with the EGS method for quickly determining the focus position within the entire focus stack. However, its focusing principle relies on the assumption that optical flow exists and remains constant, which does not hold in most focusing scenarios.
An in-depth investigation of event-driven AF in \cite{bao2023improving} revealed that pixel brightness changes during focusing exhibit symmetry around the focus position. This insight led to the development of the PBF algorithm, which identifies the focus position by locating the center of symmetry in the event polarity rate (EPR) across the focus stack. Experiments demonstrated the PBF algorithm's robustness in complex scenarios, including both dynamic and static scenes, as well as challenging conditions such as strobe flashes.

Both EGS and PSF, as well as other event-driven methods such as Ge’s~\cite{ge2023millisecond} and Lou’s~\cite{lou2023all}, require capturing a complete focus stack, spanning from defocus to near-focus and back to defocus. They analyze the entire stack to locate the \fp and drive the motor to the target position, a process that is often too slow and results in a poor user experience due to ``focus hunting''. Additionally, the focus evaluation functions they use are ``peaked'' functions, making them vulnerable to multiple peaks in the presence of interference, which can lead to suboptimal focusing. The most critical limitation of EGS and PBF is their inability to predict in real time whether the focus motor is moving toward the \fp, which can result in misguided ``focus hunting'', further increasing the focusing time.

\section{Proposed method}
Our goal is to develop a one-step event-driven AF method that can predict in real time whether to move toward the focus position and promptly signal the focus motor to stop once the focus position is achieved. Compared to existing event-driven AF methods, this approach reduces focusing time by up to two-thirds, including motor runtime. Furthermore, our one-step AF approach eliminates ``focus hunting'', significantly streamlining the focusing process.
\subsection{Background and Basics}
\PAR{Event camera.}~Each pixel on the event cameras can respond to changes in the brightness $L(x,y,t)$.
\cref{dvs_principle} shows how the event camera works:
\begin{equation}
\label{dvs_principle}
p{(x,y,t)}=\left\{
\begin{aligned}
+1 & ,\quad if& \Delta L>C,\\
-1 & ,\quad if& \Delta L<-C,
\end{aligned}
\right.
\end{equation}
where $\Delta L=L(x,y,t_i)-L(x,y,t_{i-1})$ denotes the change in brightness since the last trigger event, $C$ denotes the contrast threshold, and $p{(x,y,t)}$ denotes the event polarity. When the brightness change exceeds the threshold, event cameras emit events with different polarities depending on the direction of the change. Brightness is a logarithmic mappping of light intensity~\cite{gallego2020event}. While events capture information about brightness changes, they do not provide exact intensity values like grayscale images.

\PAR{Event-driven Temporal-mapping Photography.}~While passively generated events do not directly represent grayscale information, prior work~\cite{bao2024temporal} has demonstrated that the timestamps of events obtained through active transmittance modulation can be mapped to grayscale values. In this method, a motorized aperture modulates the incoming light, allowing the event camera to capture the Initial Positive Event (IPE) for each pixel as the aperture opens. By mapping  IPE timestamps to grayscale values, this method enables accurate high-dynamic-range (HDR) grayscale imaging. The detailed mapping relationship is provided in Eq.~(4) of \cite{bao2024temporal}.
Event-driven Temporal-Mapping photography (EvTemMap) provides valuable grayscale information for event-only cameras, which can be utilized in event-driven one-step AF.

\subsection{Principle}
\label{sec:Principle}
\PAR{Principle of focusing optics.}~For a thin lens, the following equation holds when the image is in focus:$\frac{1}{u}+\frac{1}{v}=\frac{1}{f}$,where $u$ represents the object distance, $f$ represents the focal length of the thin lens, and $v$ represents the ideal image distance. Ideally, the point spread function (PSF) of an optical system in defocus can be modeled  as a Gaussian function of the amount of defocus $\Delta v$~\cite{bao2023improving}:
$h(\boldsymbol{x},\Delta v)=\frac{1}{\sqrt{2\pi} \sigma}\exp(-\frac{\boldsymbol{x}^2}{2\sigma^2}), \sigma=k\frac{\left| \Delta v \right|}{2v_0}D$,
where $k$ represents the inverse of the sensor pixel size, while $v_0$ and $D$ represent the ideal image distance and the exit pupil diameter, respectively. And $\sigma$ can be interpreted as the blur kernel size in pixels. In a real optical system, the PSFs are slightly distorted by aberrations, but their Root Mean Square (RMS) radius always increases with $\Delta v$. In a focusing task, the focus motor typically adjusts the image distance $v$ to achieve the position that minimizes $\left| \Delta v \right|$, \ie, the point where the blur kernel radius is minimized.

\PAR{Derivation of image differentiation during the focusing process.}~Let \( F(\boldsymbol{x}, t) \) represent a clear dynamic scene and \( h(\boldsymbol{x}, t) \) a Gaussian kernel with its variance $\sigma^2$ changing over time during focusing. The image on the image plane of an event camera, \( G(\boldsymbol{x}, t) \), is expressed as: $G(\boldsymbol{x}, t) = F(\boldsymbol{x}, t) * h(\boldsymbol{x}, t) $, where $*$ represents convolution.

We can compute the first-order derivative of $G(\boldsymbol{x}, t)$ with respect to time as follows:
\begin{equation}
\label{defocus_img_1_dt}
\frac{\partial}{\partial t} G(\boldsymbol{x}, t) =  \left(\frac{\partial}{\partial t} F(\boldsymbol{x}, t) \right) * h(\boldsymbol{x}, t) + F(\boldsymbol{x}, t) *  \left(\frac{\partial}{\partial t} h(\boldsymbol{x}, t) \right),
\end{equation}
where the first term corresponds to the temporal scene variation and the second term is associated with the focusing process. In the focusing task, the change in $F(t)$ is significantly slower than that in $h(t)$. Therefore, in our derivation, we assume $\frac{\partial}{\partial t} F(\boldsymbol{x}, t)=0$.
For simplicity, assume that the Gaussian variance satisfies $\sigma(t)^2 = \sigma_0^2 +\alpha t $. The expression in \cref{defocus_img_1_dt} can then be articulated as:
\begin{equation} 
\label{dh_dt}
\frac{\partial h(\boldsymbol{x}, t)}{\partial t} = \frac{\alpha}{2} h(\boldsymbol{x}, t) \left( \frac{\boldsymbol{x}^2}{\sigma(t)^4} - \frac{1}{\sigma(t)^2} \right).
\end{equation}
Similarly, we can compute the spatial second-order derivative of $G(\boldsymbol{x}, t)$. Due to the interchangeability of convolution and differentiation operations, $F(\boldsymbol{x}, t)*\frac{\partial^2 h(\boldsymbol{x}, t)}{\partial \boldsymbol{x}^2}= \frac{\partial^2 F(\boldsymbol{x}, t)}{\partial \boldsymbol{x}^2} * h(\boldsymbol{x}, t)$. Therefore, 
\begin{equation}
\label{dG2_dx2}
\frac{\partial^2 G(\boldsymbol{x}, t)}{\partial \boldsymbol{x}^2} =2F(\boldsymbol{x}, t) * \left( \frac{\partial^2 h(\boldsymbol{x}, t)}{\partial \boldsymbol{x}^2} \right),
\end{equation}
where
\begin{equation}
\label{dh2_dx2}
\frac{\partial^2 h(\boldsymbol{x}, t)}{\partial \boldsymbol{x}^2} = h(\boldsymbol{x}, t) \left( \frac{\boldsymbol{x}^2}{\sigma(t)^4} - \frac{1}{\sigma(t)^2} \right).
\end{equation}
We can derive $S(t)$ from \cref{defocus_img_1_dt} and \cref{dG2_dx2}:
\begin{equation}
\label{eq:st}
\begin{split}
&S(t) = -\int \frac{\partial G(\boldsymbol{x}, t)}{\partial t} 
\cdot \frac{\partial^2 G(\boldsymbol{x}, t)}{\partial \boldsymbol{x}^2} \, d\boldsymbol{x} \\
&= -\alpha \bigg[ \int \left(F(\boldsymbol{x}, t) * \frac{\partial^2 h(\boldsymbol{x}, t)}{\partial \boldsymbol{x}^2} \right)^2 \, d\boldsymbol{x} \bigg].
\end{split}
\end{equation}
The value within [] is non negative, and the sign of $S(t)$ is entirely determined by $\alpha$. 

\subsection{Event Laplacian Product}
\label{sec:ELP}

We use ``Event Laplacian Product'' (ELP) as the focus detection function for the event-driven one-step AF, which is defined as:
\begin{equation}
\label{eq:ELP}
\text{ELP}(t) = -\sum \left(\nabla^2 I(t) \cdot E(t) \right),
\end{equation}
where $I(t)$ denotes a grayscale image closest to the current time $t$, $\nabla^2$ represents the Laplacian for $I(t)$, and $E(t)$ refers to the event frame acquired at $t$. 
The definition of an event frame is as follows.
Let the event stream from an event camera be denoted by \( \{e_i\} \), where each event \( e_i \) is represented by the tuple \( (x_i, y_i, t_i, p_i) \). Here, \( (x_i, y_i) \) denotes the pixel location, \( t_i \) is the timestamp, and \( p_i \) represents the event polarity. The event frame \( E(x, y, t-\Delta t, t) \), which accumulates events over the time interval \( [t-\Delta t, t] \), is mathematically expressed as:
\begin{equation}
\label{eq:event_frame}
E(x, y, t-\Delta t, t) = \sum_{i} p_i \cdot \delta(x - x_i, y - y_i).
\end{equation}
$\text{ELP}(t)$ in \cref{eq:ELP} is the difference approximation of $S(t)$ in \cref{eq:st}, where $E(t)\approx \frac{\partial}{\partial t} G(\boldsymbol{x}, t)$ and $I(t)\approx G(t)$. The gap between $G(t)$ and $I(t)$ is essentially whether $h$ is the current defocus kernel. According to \cref{eq:st}, $h(\boldsymbol{x},t)$ can be replaced with $h(\boldsymbol{x},t')$ at any given moment $t'$ without affecting the sign of $S(t)$, as well as $\text{ELP}(t)$. 
The gap between $E(t)$ and $\frac{\partial}{\partial t} G(\boldsymbol{x}, t)$ results in fluctuations in ELP values.

\Cref{fig:ELP} illustrates how the ELP varies with \( \Delta v \) as the system transitions from defocus to near-focus and back to defocus. As the focusing process approaches the focus position, the ELP value increases steadily. Upon reaching the focus position, the ELP undergoes a sharp ``sign mutation'' from positive to negative.
Comparing event frames with the Laplacian of grayscale images at different \( \Delta v \) reveals the underlying causes of ELP value changes. As the system moves closer to the focus position, the coefficient \( \alpha \) in \cref{dh_dt} is negative, causing negative events to occur in regions where the Laplacian is positive, thereby resulting in a positive ELP value.  When \( \left| \Delta v \right| \) is large, the Laplacian values are generally low, and events are dispersed, resulting in a low ELP value. Approaching the focus position, Laplacian values increase, and the event frames sharpen progressively, leading to a surge in the absolute ELP value. Upon crossing the focus position, \( \alpha \) in \cref{dh_dt} turns positive, resulting in positive events occurring in regions of positive Laplacian, which induces an ELP ``sign mutation''. 

The combination of the grayscale Laplacian and events gives ELP excellent properties as a focus detection function: (1) Its positive and negative values indicate whether the system is moving toward or away from the focus position, helping to avoid misguided ``focus hunting''; (2) ELP is highly sensitive in identifying the focus position, with the abrupt ``sign mutation'' typically occurring within a single depth of focus; (3) ELP does not require \textbf{repeated searches} for the focus position, but only needs to \textbf{detect} the ``sign mutation'' \textbf{once}, making it a one-step AF solution.

\subsection{ELP adaptive filter}
\label{sec:elp_filter}
In \cref{eq:event_frame}, the time interval $\Delta t$ of event frames can be made very short ($<$1ms) to enhance the focus sensitivity. However, if $\Delta t$ is too short, focus events may become susceptible to noise, leading to local fluctuations in the ELP. 
To address this, we introduce an adaptive ELP filter to suppress fluctuations. The filter is defined as follows:

1. Calculate the average of the past \( W \) collected ELP values, \( \overline{ELP} = \frac{1}{W} \sum_{i=1}^{W} ELP_{i} \).

2. If \( \left| ELP_\text{now} - \overline{ELP} \right| < ELP_\text{thd} \), return the filtered value \( ELP_\text{filtered} = S \cdot ELP_\text{now} + (1 - S) \cdot \overline{ELP} \); otherwise, return the original value \( ELP_\text{filtered} = ELP_\text{now} \).

The ELP adaptive filter determines whether to apply filtering based on the comparative judgment, which preserves the steepness of the ELP at the ``sign mutation'' while smoothing out local fluctuations elsewhere. The smoothing factor \( S \), where \( S \in [0, 1] \), controls the level of smoothness, with smaller values of $S$ resulting in greater smoothness. The window size \( W \) defines the filter's visible range.
The black solid line in \cref{fig:ELP} shows the filtered ELP, which removes local fluctuations in the green dashed line while preserving the steepness at the ``sign mutation'' point.

\subsection{Event-only one-step AF pipeline}
\label{sec:pipeline}
\begin{figure}[t]
  \centering
   \includegraphics[width=0.99\linewidth]{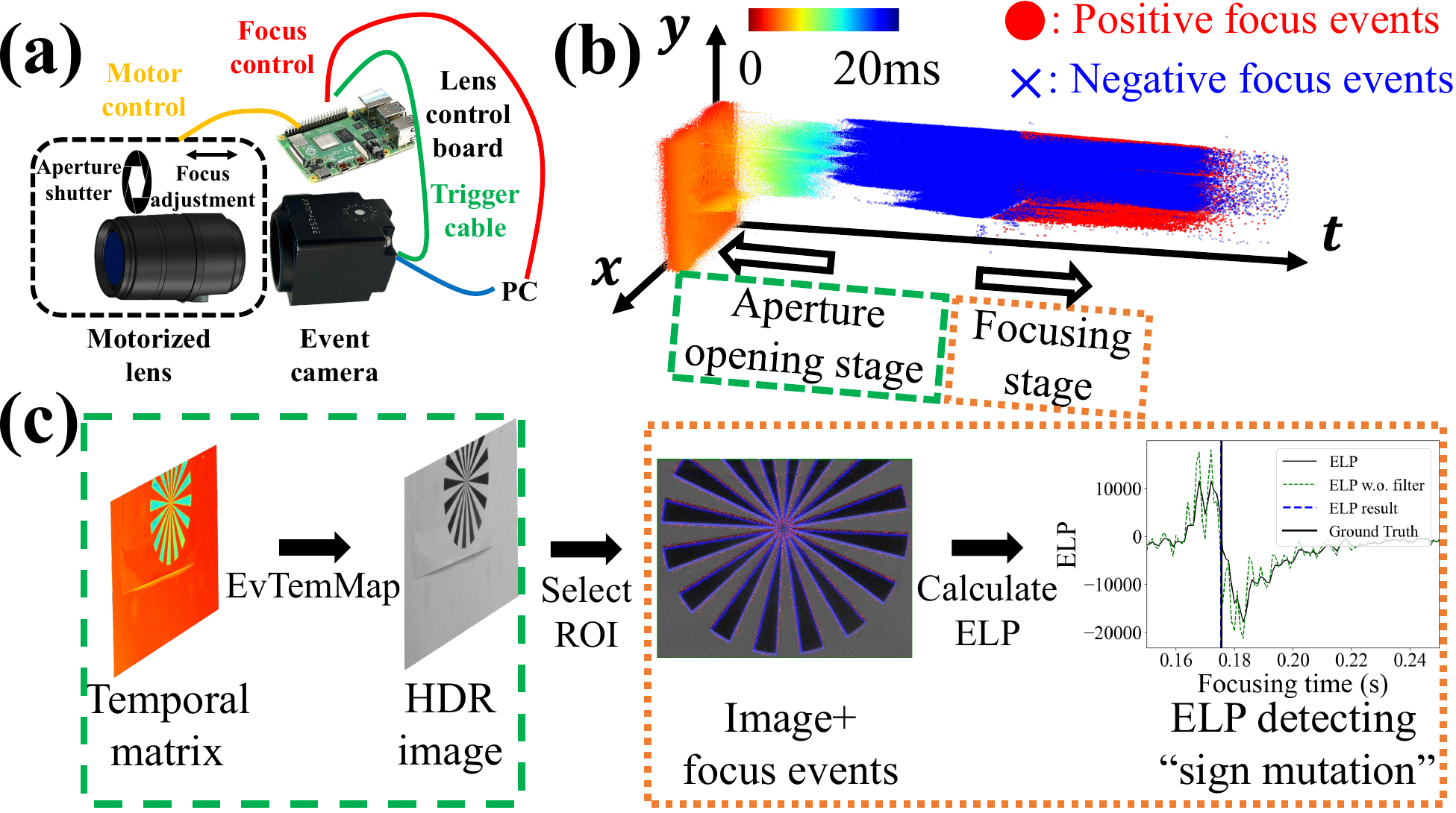}
   \caption{Event-driven one-step AF setup and pipeline. (a) Hardware setup. (b) Collected real events. (c) The pipeline of grayscale image acquisition and ELP value calculation.}
   \label{fig:pipeline}
\end{figure}

The event-only one-step AF pipeline consists of two stages: (1) \textbf{Aperture opening stage}: EvTemMap~\cite{bao2024temporal} is applied to capture a grayscale image; (2) \textbf{Focusing stage}: After selecting the focus region of interest (ROI), focus events are captured to compute ELP values and detect ``sign mutation''.

\Cref{fig:pipeline}~(a) shows the ELP hardware setup. \Cref{fig:pipeline}~(b) shows real events captured with a Prophesee EVK4 (an event-only camera), with 20 ms for the aperture opening and 100 ms for the focusing stage. The IPEs from aperture opening stage are color-coded by timestamp, while focus events are marked with blue `x' for negative events and red `o' for positive events. To illustrate the entire variation in ELP, we also capture focus events during the additional defocusing process. 
As shown in \Cref{fig:pipeline}~(c), in the aperture opening stage, following the principle of EvTemMap~\cite{bao2024temporal}, we capture only the IPE from each pixel, forming a ``Temporal matrix''. After real-time EvTemMap, the ``Temporal matrix'' is converted into an HDR grayscale image, allowing the user to select the focus ROI. The focus motor then begins to move, and the event camera continuously captures focus events, which are combined with the previously obtained grayscale image to compute ELP values. 
During the focusing stage, the AF system continuously monitors changes in ELP: (1) If the ELP value is negative, the motor reverses direction toward the focus position; (2) When the ELP value shows an abrupt ``sign mutation'', indicating the focus position has been reached, the motor stops.

Notably, event cameras like the DAVIS346 can capture grayscale images directly and output them in sync with events, eliminating the need for an aperture opening stage.
\section{Experiment}
\label{sec:experiment}
We compare the proposed ELP method with two event-driven AF methods EGS~\cite{lin2022autofocus} and PBF~\cite{bao2023improving} on synthetic datasets as well as the DAVIS346 and Prophesee EVK4 datasets, demonstrating the accuracy and efficiency of ELP.
\subsection{Synthetic experiment setup}
\label{sec:syn_exp}
\begin{figure}[t]
  \centering
   \includegraphics[width=0.99\linewidth]{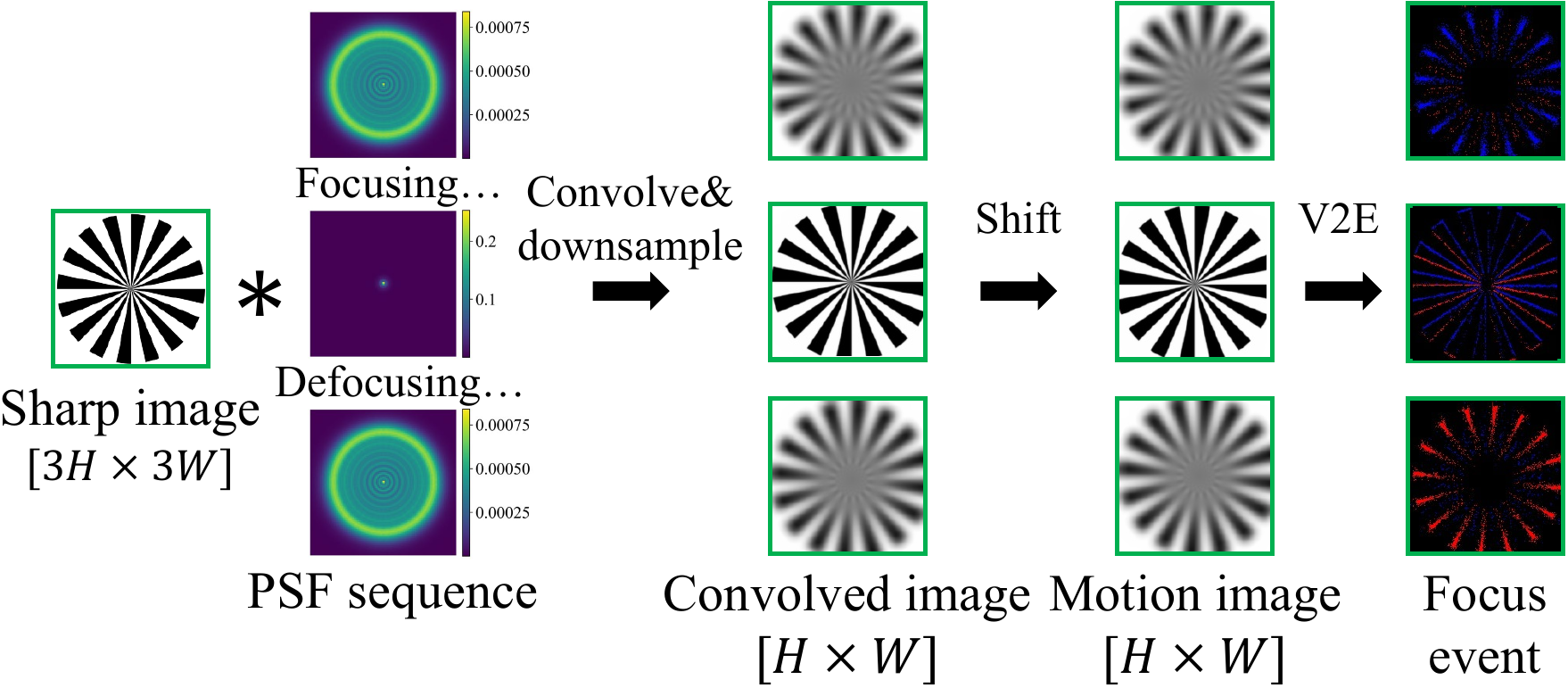}
   \caption{The pipeline of PSF-based focus event simulator.}
   \label{fig:syn}
\end{figure}
We develop a PSF-based focus event simulator to accurately generate focus events of real lenses, since the blur kernels of a real lens during focusing are determined by the PSFs at different $\Delta v$, which, due to lens aberrations, differ from ideal Gaussian blur kernels.

\PAR{Simulator Overview.}~\Cref{fig:syn} illustrates the pipeline for generating synthetic PSF-based focus events. The process begins by loading a sharp grayscale image with a resolution of \([3H \times 3W]\). This image is then convolved with \( psf(\Delta v) \) at each defocus position \( \Delta v \) to generate the corresponding image, which is subsequently downsampled to a resolution of \([H \times W]\). In the synthetic dataset, $H$ and $W$ are equal to 200 pixels. To simulate motion during focusing, we translate the convolved image before generating events using the event simulator (``V2E'')~\cite{hu2021v2e}. The focus event stack spans 1 second: ELP utilizes only the first 0.5 seconds of events (focusing process), while EGS and PBF use the entire 1-second focus event stack (focusing and defocusing process).

\PAR{PSF characteristics.}~The synthetic dataset includes four groups of PSF sequences, corresponding to four fields of view (FoV), with each sequence containing 1001 PSFs sampled uniformly within the range of $\Delta v \in [-400,400]$ $\mu$m. The RMS radius of the PSF at maximum defocus is 10 times that at the focus position, indicating a challenging initial defocus.
For further details on the PSF, please refer to the supplementary material.

\PAR{Motion simulation.}~The motion vector, $\mathbf{t}$, for each frame consists of two components: (1) random jitter $v_{\text{jitter}}$ and (2) constant motion $v_{\text{motion}}$, expressed as $\mathbf{t} \sim \mathcal{N}(v_{\text{motion}}, v_{\text{jitter}}^2)$.  In the synthetic dataset, we provide two types of motion parameters: moderate and violent. For the moderate motion, $v_{\text{motion}}=[3,3]$ pixels/s and $v_{\text{jitter}}=[20,20]$ pixels/s, whereas for the violent motion, $v_{\text{motion}}=[3,3]$ pixels/s and $v_{\text{jitter}}=[100,100]$ pixels/s.

\PAR{Brightness simulation.}~Brightness affects the frame rate of grayscale image capture, impacting the acquisition of the Laplacian in the ELP method. Under normal brightness ($>$100 lux), the DAVIS346 camera captures images at up to 50 frames per second (FPS). However, at lower brightness ($<$1 lux), the grayscale image frame rate drops to 20 FPS or lower. We evaluate the performance of ELP at both 50 FPS and 20 FPS. Additionally, since the event-only one-step AF system mentioned in \cref{sec:pipeline} uses only a single defocus image, we also simulate this scenario.

\subsection{Synthetic experiment result }
\begin{table}[h!]
\small
\centering
\renewcommand{\arraystretch}{0.7} 
\setlength{\tabcolsep}{6pt} 
\begin{tabular}{>{\centering\arraybackslash}p{1.6cm} >{\centering\arraybackslash}p{1.5cm} >{\centering\arraybackslash}p{1.5cm} >{\centering\arraybackslash}p{1.5cm}}
\toprule
\multirow{2}{*}{\textbf{Method}} & \multicolumn{3}{c}{\textbf{Motion state}} \\
\cmidrule(lr){2-4}
 & \textbf{Static} & \textbf{Moderate} & \textbf{Violent} \\
\midrule
EGS & 33.31  & 29.33 & 21.78 \\
PBF & 4.93  & 3.99  & 2.69  \\
ELP(1FPS) & \textbf{2.00}  & 3.66 & 3.66 \\
ELP(20FPS) & 2.00 & 2.40 & 2.97 \\
ELP(50FPS) & 2.00 & \textbf{1.51} & \textbf{2.26} \\
\bottomrule
\end{tabular}
\caption{MAE comparison on synthetic datasets (in $\mu$m). ELP method is tested with various grayscale frame rates under three lighting conditions. The best performances are marked in \textbf{bold}.}
\label{table:syn}
\end{table}
In the synthetic dataset of 84 cases, the EGS method yields the highest mean absolute error (MAE), as shown in \cref{table:syn}. In contrast, both the PBF and ELP methods achieve near-optimal results across various motion states. 
Detailed results for each case in the synthetic dataset are provided in the supplementary material, where the focusing errors for both PBF and ELP remain within one depth of focus (16~$\mu$m) for each case. However, the EGS method fails to provide a focus position in 28.6\% of cases, and in 68.3\% of the remaining cases, its error exceeds one depth of focus.
With its one-step AF principle, ELP eliminates the need for defocusing and returning to the target, reducing focusing time by two-thirds compared to PBF and EGS.
\begin{figure}[ht]
  \centering
   \includegraphics[width=0.99\linewidth]{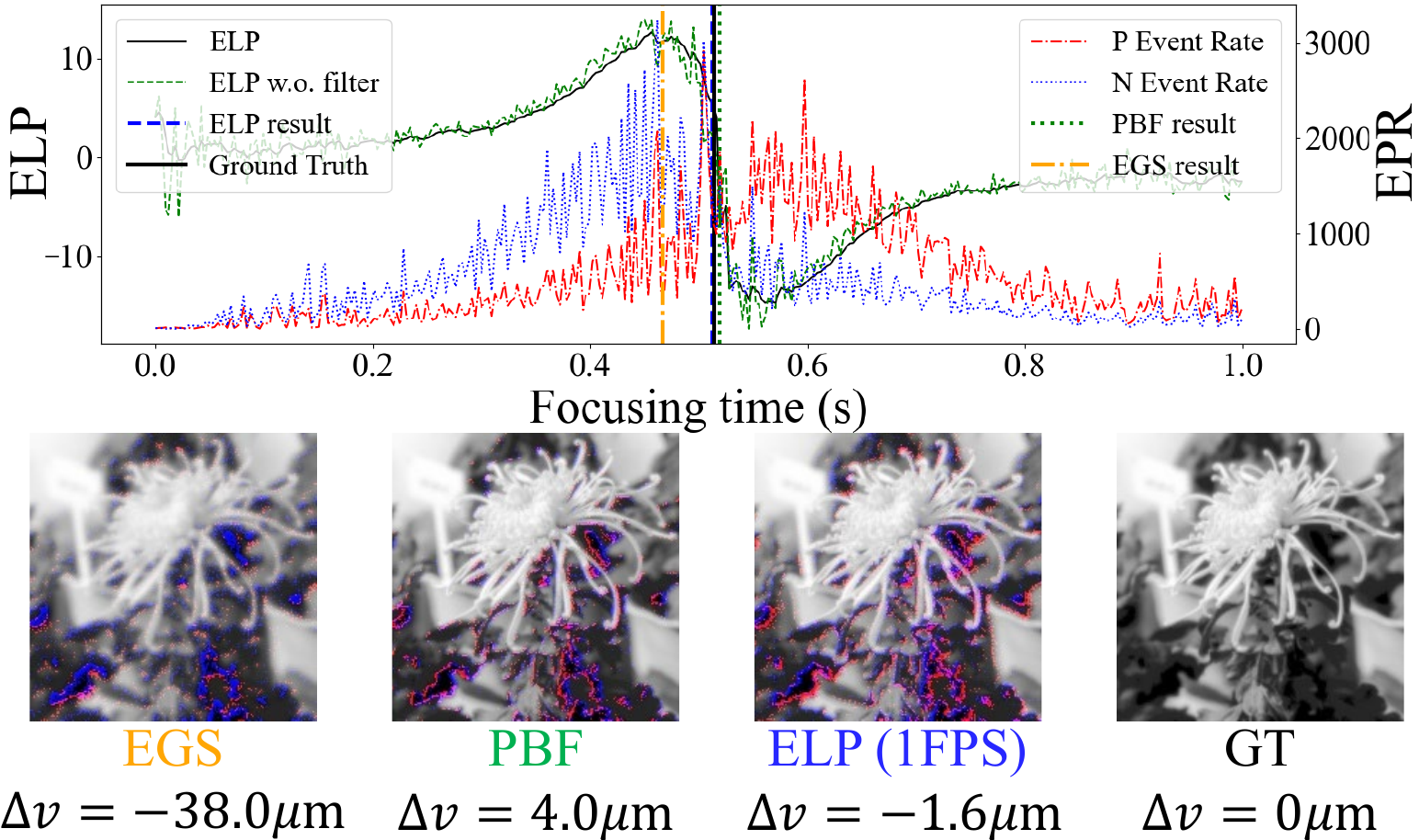}
   \caption{Visualization of a violent shake scene. Top row: temporal variation of ELP and EPR across the entire focus stack. Bottom row: grayscale images at different $\Delta v$ overlaid with focus events.}
   \label{fig:syn_cat}
\end{figure}

\Cref{fig:syn_cat} presents a visual comparison of three event-driven AF methods in a violent motion scenario, where ELP utilizes only one grayscale frame (1 FPS) for Laplacian information. In this case, both PBF and ELP achieve sharp focus, while the blur kernel of the EGS method exceeds two pixels, as shown in the bottom row. The top row reveals that the severe jitter causes significant localized fluctuations in both the EPR and the ELP curves. The ELP adaptive filter effectively suppresses these fluctuations while preserving the steepness of the ``sign mutation'' point, ensuring the accuracy of ELP. In instances of random violent jitter, EGS incorrectly identifies the location with the highest ER as the focus position, whereas PBF accurately locates the center of symmetry of the EPR, closely aligning with the Ground Truth (GT) focus position.

\subsection{Real experiment setup}
\PAR{Focusing hardware and configuration.}~In real experiments, a motorized focusing lens is used for the focusing process, with DAVIS346 and Prophesee EVK4 cameras capturing events (and, if available, grayscale images). The focus ring of the motorized lens is driven by a stepper motor. When the lens reaches the GT \fp, a trigger signal is sent to the AF control unit to mark the GT timestamp during focusing. The GT focus position is determined by locating the point where the blinking focus star appears sharpest. The complete focus event stack—from defocus to near-focus and back to defocus—spans approximately 200 ms, starting from a significant $\Delta v$ of 200 $\mu$m.

\PAR{DAVIS dataset.}~In the DAVIS dataset, time-synchronized event and grayscale images are captured simultaneously. The maximum frame rate for grayscale images is 50 FPS, but it can drop to as low as 20 FPS under low-light conditions. ELP uses real-time events and the latest  grayscale frame to compute the value and detects in real time whether the ``sign mutation'' has happened. In contrast, EGS and PBF require capturing the entire focus event stack, performing an algorithmic search for the focus position, and then driving the motor to that position.

\PAR{EVK4 dataset.}~Since EVK4 only provides events, we build a \textbf{event-only one-step AF system} (\cf~\cref{sec:pipeline}). ELP computes the focus detection function using a single frame of the defocus image acquired from the aperture opening stage, along with the subsequent real-time focus events. In contrast, EGS and PBF do not use events from the aperture opening stage; instead, they rely on the event stack from the entire focusing stage. The aperture is opened by the motorized lens's high-speed motor, transitioning from fully closed to fully open within 20 ms.

\subsection{Real experiment result}
\label{sec:real_result}
\PAR{One-step AF visualization.}~The supplementary videos provide a visualization of the one-step AF process of the ELP method on real datasets, showcasing its ability to drive directly to the focus position with precision and efficiency.

\begin{table*}[htbp]
\centering
\renewcommand{\arraystretch}{0.7} 
\small 
\begin{tabular}{p{0.9cm} p{1.4cm} p{0.75cm} p{0.75cm} p{0.75cm} p{0.75cm} p{0.75cm} p{0.75cm} p{0.75cm} p{0.75cm} p{0.75cm} p{0.75cm} p{0.75cm} p{0.75cm}}
\toprule
\multirow{2}{*}{dataset} & \multirow{2}{*}{scene} & \multicolumn{3}{c}{bright\_dynamic} & \multicolumn{3}{c}{bright\_static} & \multicolumn{3}{c}{dark\_dynamic} & \multicolumn{3}{c}{dark\_static} \\
\cmidrule(lr){3-5} \cmidrule(lr){6-8} \cmidrule(lr){9-11} \cmidrule(lr){12-14}
 & & EGS & PBF & ELP & EGS & PBF & ELP & EGS & PBF & ELP & EGS & PBF & ELP \\
\midrule
\multirow{7}{*}{DAVIS} & box & -10.8 & -1.4 & \textbf{0.3} & -33.1 & -9.2 & \textbf{0.2} & -27.1 & \textbf{0.1} & 0.5 & -25.4 & -7.2 & \textbf{0.2} \\
 & focus star & -26.7 & 2.4 & \textbf{0.4} & -44.6 & -8.0 & \textbf{0.0} & -28.9 & -3.0 & \textbf{0.0} & -27.2 & -3.4 & \textbf{0.7} \\
 & forest & -12.1 & -3.1 & \textbf{0.6} & -14.8 & -6.4 & \textbf{-0.7} & -11.5 & 1.4 & \textbf{0.2} & -13.1 & -5.9 & \textbf{0.2} \\
 & ghost & -21.2 & -19.4 & \textbf{0.6} & -54.1 & -25.6 & \textbf{0.4} & -24.3 & -14.1 & \textbf{-0.1} & -24.9 & -7.7 & \textbf{0.3} \\
 & lens & -44.5 & 4.3 & \textbf{0.7} & -42.3 & 3.4 & \textbf{-0.2} & -39.1 & -35.5 & \textbf{0.2} & -37.8 & 6.7 & \textbf{0.0} \\
 & mountain & -12.4 & 1.1 & \textbf{0.1} & -28.4 & -7.2 & \textbf{-0.4} & -13.6 & 2.1 & \textbf{0.2} & -22.3 & -5.0 & \textbf{0.0} \\
 & statue & -65.4 & 9.1 & \textbf{0.1} & -38.7 & -16.8 & \textbf{0.2} & -26.4 & -13.2 & \textbf{0.2} & -25.1 & -1.5 & \textbf{0.6} \\
\midrule
\multirow{7}{*}{EVK4} & box & -17.4 & 10.6 & \textbf{0.6} & -32.4 & -1.9 & \textbf{0.1} & -15.6 & 13.7 & \textbf{-0.3} & -26.2 & 4.7 & \textbf{-1.3} \\
 & focus star & 29.8 & -6.6 & \textbf{0.4} & -26.0 & 4.0 & \textbf{0.0} & 86.9 & 5.9 & \textbf{-0.1} & 88.1 & 10.1 & \textbf{0.1} \\
 & forest & -16.3 & -3.7 & \textbf{0.3} & -9.1 & 6.1 & \textbf{0.1} & -9.4 & 9.7 & \textbf{-0.3} & -10.8 & 6.8 & \textbf{-0.2} \\
 & ghost & -38.4 & 18.4 & \textbf{0.4} & -37.5 & \textbf{0.1} & \textbf{0.1} & -6.2 & 35.3 & \textbf{-0.7} & -16.5 & 7.2 & \textbf{3.2} \\
 & lens & -26.4 & -4.3 & \textbf{0.7} & 126.3 & -5.8 & \textbf{0.2} & 120.5 & 19.3 & \textbf{1.3} & 164.8 & 72.7 & \textbf{-0.3} \\
 & mountain & -20.2 & -2.3 & \textbf{-0.3} & -18.4 & 2.0 & \textbf{0.0} & -15.4 & 7.7 & \textbf{-0.3} & -44.2 & 40.4 & \textbf{0.4} \\
 & statue & -36.3 & 10.7 & \textbf{0.7} & -37.3 & 7.0 & \textbf{0.0} & -48.4 & -7.3 & \textbf{1.7} & -26.5 & 26.8 & \textbf{1.8} \\
\bottomrule
\end{tabular}
\caption{Quantitative comparison of focusing errors across event-driven AF methods on each case of real datasets, measured in $\mu$m. The best performances are marked in \textbf{bold}.}
\label{tab:real_quan}
\end{table*}

\PAR{Quantitative results.}~\Cref{tab:real_quan} details the quantitative focusing errors of the EGS, PBF, and our ELP methods on the real datasets. The motorized focusing len is considered accurately focused if the focusing error is within 6 $\mu$m (\ie one depth of focus). The ELP method achieves accurate focus in every case for both the DAVIS and EVK4 datasets, while the PBF method achieves accurate focus in approximately 43\% of the cases, and the EGS method fails to achieve accurate focus in any case. \Cref{tab:real_summ} summarizes the MAE of the three event-driven AF methods on both datasets, showing that the focusing error of the ELP method is reduced by 24 times compared to the state-of-the-art PBF method on the DAVIS346 dataset and by 22 times on the EVK4 dataset. 

\begin{table}[htbp]
\centering
\renewcommand{\arraystretch}{0.7} 
\small
\begin{tabular}{c@{\hskip 2.5pt}c@{\hskip 2.5pt}c@{\hskip 2.5pt}c@{\hskip 2.5pt}c}
\toprule
Dataset & Method & MAE ($\mu$m) & Runtime (ms) & Focusing time (ms) \\
\midrule
\multirow{3}{*}{DAVIS} & EGS & 28.42 & 10.81 & 310.81 \\
                       & PBF & 7.02 & \textbf{2.37} & 302.37 \\
                       & ELP & \textbf{0.29} & 12.14 & \textbf{103.36} \\
\midrule
\multirow{3}{*}{EVK4}  & EGS & 41.12 & 104.25 & 404.25 \\
                       & PBF & 12.54 & \textbf{2.52} & 302.52 \\
                       & ELP & \textbf{0.57} & 83.00 & \textbf{112.97} \\
\bottomrule
\end{tabular}
\caption{Quantitative comparison of focusing performance. The best performances are marked in \textbf{bold}.}
\label{tab:real_summ}
\end{table}

\PAR{Speed analysis.}~The focusing speed analysis for the three event-driven AF methods is summarized in \cref{tab:real_summ}, where ``Runtime'' refers to the algorithm's execution time, and ``Focusing time'' includes the entire focusing process, encompassing motor operation and data acquisition. 
In terms of algorithm runtime, the PBF algorithm performs best on both datasets, with a complexity of \( O\left( \frac{1}{\Delta t} \right) \)~\cite{bao2023improving}, where $1/\Delta t$ is the sampling rate of event frames. The EGS algorithm performs well on the DAVIS dataset but worst on the EVK4 dataset, due to its complexity of \(O(N_e)\)~\cite{lin2022autofocus}, where $N_e$ denotes the ER. Our ELP algorithm demonstrates moderate runtime performance on both datasets, with a complexity of \( O\left( \frac{k}{\Delta t} \right) \), where \( k \) represents the number of non-zero pixels in the event frame.
For the total time required for the complete focusing process, however, our ELP method is nearly three times faster than the other two methods. This efficiency results from ELP's one-step AF principle, which continuously detects in real time if the \fp is reached as the focus motor moves, rather than waiting for the full focusing trip to complete and then analyzing the entire focus stack before positioning the motor.

\begin{figure*}[ht]
  \centering
   \includegraphics[width=0.99\linewidth]{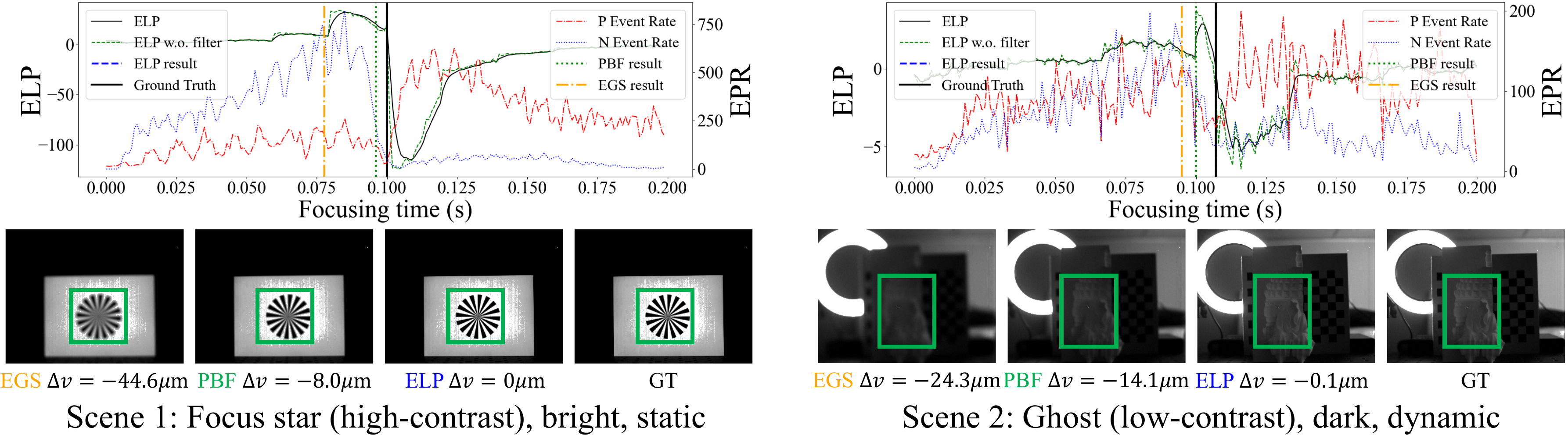}
   \caption{Visualization of two scenes from the DAVIS dataset. Top row: temporal variation of ELP and EPR across the entire focus stack. Bottom row: grayscale images corresponding to the focus positions determined by the three event-driven methods, alongside the GT. Green boxes indicate the focus ROI.}
   \label{fig:davis}
\end{figure*}

\PAR{Impact of contrast, brightness, and motion.}~A high-contrast, bright, and static scene is a straightforward case for the focusing task, as seen in Scene 1 of \cref{fig:davis}. In this scene, ELP achieves precise focus, whereas PBF exhibits a focusing error exceeding one depth of focus, and EGS shows an even larger error, exceeding seven depths of focus. Observing the changes in ELP and EPR over time, the ``sign mutation'' point of the ELP curve is very close to the symmetric position of the positive and negative EPR curves. Recalling \cref{eq:event_frame}, ELP, like PBF, also incorporates event polarity information. The \textbf{search} for the symmetry center of EPR in PBF is transformed into the \textbf{detection} of the ``sign mutation'' in ELP using the additional spatial texture information provided by the Laplacian, enabling one-step AF. In the absence of optical flow, the principle of EGS cannot hold, which explains its failure.

In a low-contrast, dark, and dynamic scene~(Scene 2 in~\cref{fig:davis}), the signal-to-noise ratio of the focus event drops sharply, resulting in significant fluctuations in both ELP and EPR curves. Under these challenging conditions, EGS demonstrates a focusing error of four depths of focus, whereas PBF exceeds two depths of focus, resulting in noticeable blurring. In contrast, ELP consistently achieves precise focus. This example highlights how incorporating Laplacian information from grayscale images enhances the event-driven AF robustness in extreme scenarios.

\begin{figure*}[ht]
  \centering
   \includegraphics[width=0.99\linewidth]{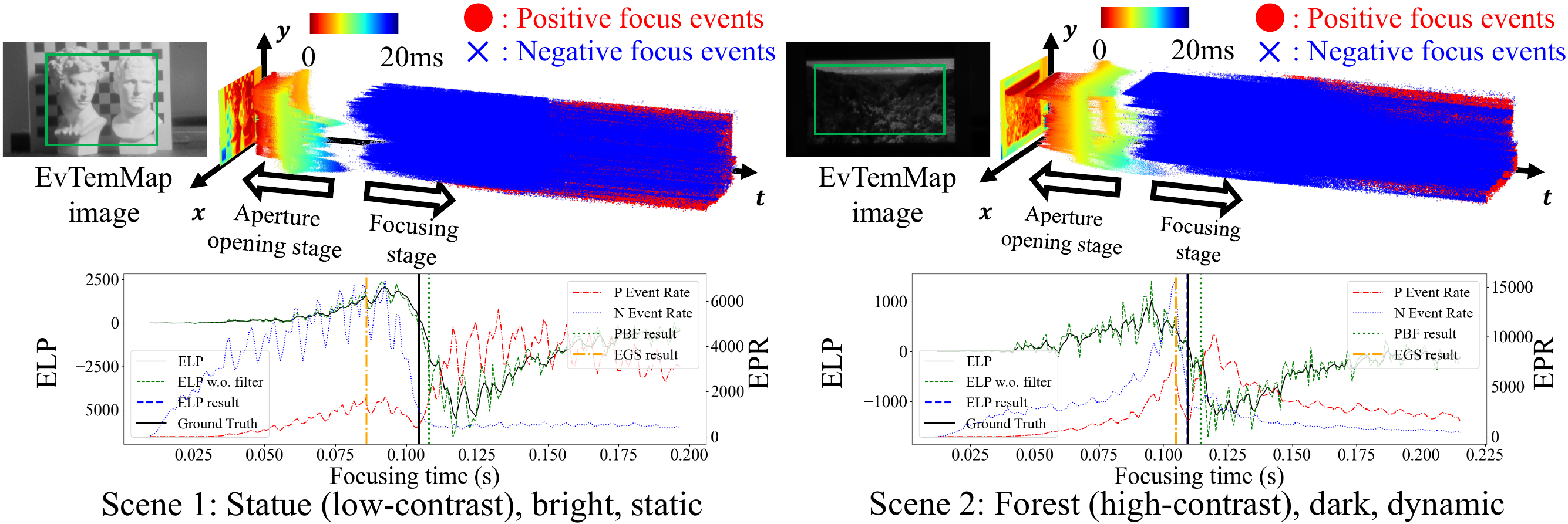}
   \caption{Visualization of two scenes from the EVK4 dataset. Top row: visualization of raw event data and initial defocus grayscale images obtained from EvTemMap~\cite{bao2024temporal}, with green boxes indicating the focus ROI. Bottom row: temporal variation of ELP and EPR.}
   \label{fig:evk4}
\end{figure*}

\PAR{Event-only one-step AF.}~\Cref{fig:evk4} shows two scenes from the EVK4 dataset captured by the event-only one-step AF system. Although the event-only system is limited to acquiring a single defocus grayscale frame, it benefits from EvTemMap's high dynamic range, ultra-high grayscale resolution, and large depth of field~\cite{bao2024temporal}. The increased dynamic range allows ELP to utilize more texture information, while the enhanced grayscale resolution improves the accuracy of Laplacian computations. Additionally, the larger depth of field enables the frame to serve as a sharp texture reference for ROIs at different focus positions, similar to an all-in-focus frame.
The objects captured in Scene 1 of \cref{fig:evk4} and Scene 2 of \cref{fig:davis} are both low-contrast plaster statues. However, the high grayscale resolution and dynamic range of EvTemMap provide a more accurate Laplacian, allowing ELP to perform effectively even with only a single frame for reference. In Scene 2 of \cref{fig:evk4}, characterized by richer textures and more motion events, the ELP curve shows greater local fluctuations. The ELP adaptive filter effectively reduces most of these fluctuations, ensuring robust results. Comparing the two scenes in \cref{fig:evk4}, the richer texture and stabilized optical flow in Scene 2 enable EGS to achieve better focusing results. Both PBF and ELP, which incorporate event polarity, consistently achieve accurate focus across different scenarios.


\begin{table}[htbp]
    \centering
    \renewcommand{\arraystretch}{0.7} 
    \small
    \begin{tabular}{cccc}
        \toprule
        Dataset & ELP & w.o. filter & w.o. Laplacian \\
        \midrule
        DAVIS & 0.29 & 2.12 & 2.48 \\
        EVK4 & 0.57 & 19.85 & 7.81 \\
        \bottomrule
    \end{tabular}
    \caption{Ablation experiment results. Metric: MAE ($\mu$m).}
    \label{tab:ablation}
\end{table}
\begin{figure}[t]
  \centering
   \includegraphics[width=0.98\linewidth]{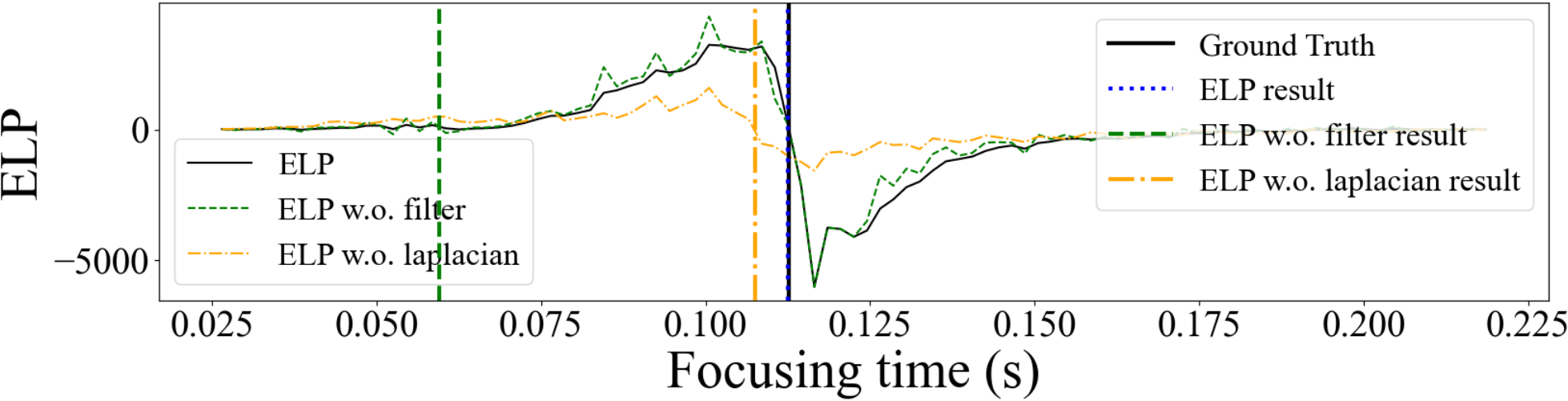}
   \caption{Visualization of ELP curves from the ablation study.}
   \label{fig:ablation}
\end{figure}
\PAR{Ablation study.}~\Cref{tab:ablation} summarizes the results of ablation experiments on real datasets, while \cref{fig:ablation} illustrates the ELP curves for a specific case (mountain, bright, dynamic) in the EVK4 dataset under two ablation settings.
In the ELP method, the adaptive filter is crucial for robustness, especially in the EVK4 dataset, which uses a single grayscale frame. As shown by the green dashed line in \cref{fig:ablation}, removing this filter introduces noise fluctuations, causing premature false focus detections. Replacing the Laplacian with a uniform 1 matrix flattens the orange dashed-dotted ELP curve, leading to an earlier ``sign mutation''.
Without Laplacian information, ELP degenerates into the difference between negative and positive event rates, losing the ability to determine the focus direction based on sign. Consequently, in 40\% of cases in the DAVIS dataset, the ELP without Laplacian provides incorrect focus adjustment direction, resulting in ``focus hunting''. 
Additionally, for the event-only EVK4 dataset, we ablate the EvTemMap method and instead use E2VID to obtain grayscale Laplacian information. Detailed results are provided in the supplementary material.



\section{Discussion} 
We introduce the first event-driven, one-step AF method, the Event Laplacian Product (ELP), which reduces focus time to one-third of existing event-driven methods and resolves the issue of ``focus hunting''. Experiments on synthetic data and two real-world event camera datasets across diverse lighting and motion conditions demonstrate that ELP consistently achieves precise focus within one depth of field. Compared to the state-of-the-art event-driven method PBF, ELP reduces focusing error by 24 times on the DAVIS346 dataset and by 22 times on the EVK4 dataset.

ELP is typically fast and accurate across most scenarios but faces challenges under extreme high-speed motion in event-only settings. Future work could focus on adaptive ``sign mutation'' detection to enhance ELP's robustness, ensuring reliable performance in demanding conditions.
{
    \small
    \bibliographystyle{ieeenat_fullname}
    \bibliography{main}
}

\clearpage
\setcounter{page}{1}
\maketitlesupplementary


\section{PSF-based focus event simulator details}
\begin{figure*}[ht]
  \centering
   \includegraphics[width=0.98\linewidth]{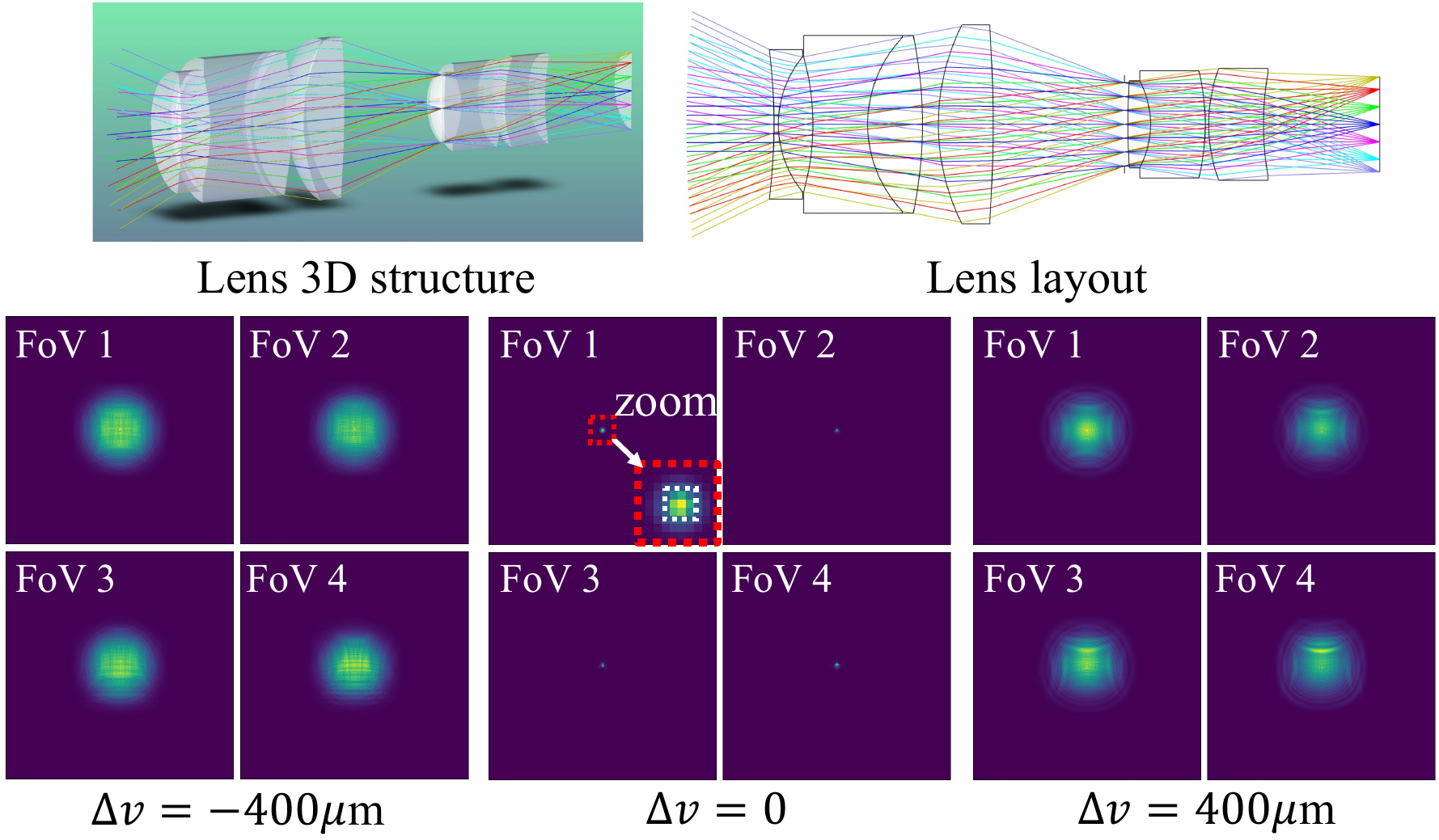}
   \caption{Details of the real 50mm F2.0 lens used for the synthetic dataset. Top row: lens structure. Bottom row: PSFs of the 4 FoVs at different $\Delta v$. At the focus position position, the RMS radius of the PSF is less than 1. Since the convolved images are 1/3 downsampled, the blur kernel is an ideal Dirac function $\delta(\boldsymbol{x})$ at the focus position.}
   \label{fig:psf_detail}
\end{figure*}

\Cref{fig:psf_detail} shows the details of the real 50mm F2.0 lens used in the synthetic dataset, where the top row shows the structure of the lens.
Of interest is the PSF of this lens at different $\Delta v$, as shown in the bottom row. 
At the focus position, the RMS radius of the PSF is less than 1, \ie, the vast majority of the energy is concentrated in the $3\times3$ region within the white dashed box. According to the paragraph \textbf{Simulator Overview} in \cref{sec:syn_exp}, the convolved image is 1/3 downsampled, so the PSF at the focus position can be considered an ideal Dirac function $\delta(\boldsymbol{x})$. 

As shown by the PSFs for $\Delta v=-400 \mu$m and $\Delta v=400 \mu$m for each FoV in \cref{fig:psf_detail}, the PSFs of the real lens differ from the ideal Gaussian blur kernel. Additionally, for a real lens, the size of the blur kernel (as measured by the RMS radius of the PSF) may not necessarily change linearly with $\Delta v$, as depicted in the top row of \cref{fig:dof}.


\section{Depth of focus}
\begin{figure*}[ht]
  \centering
   \includegraphics[width=0.98\linewidth]{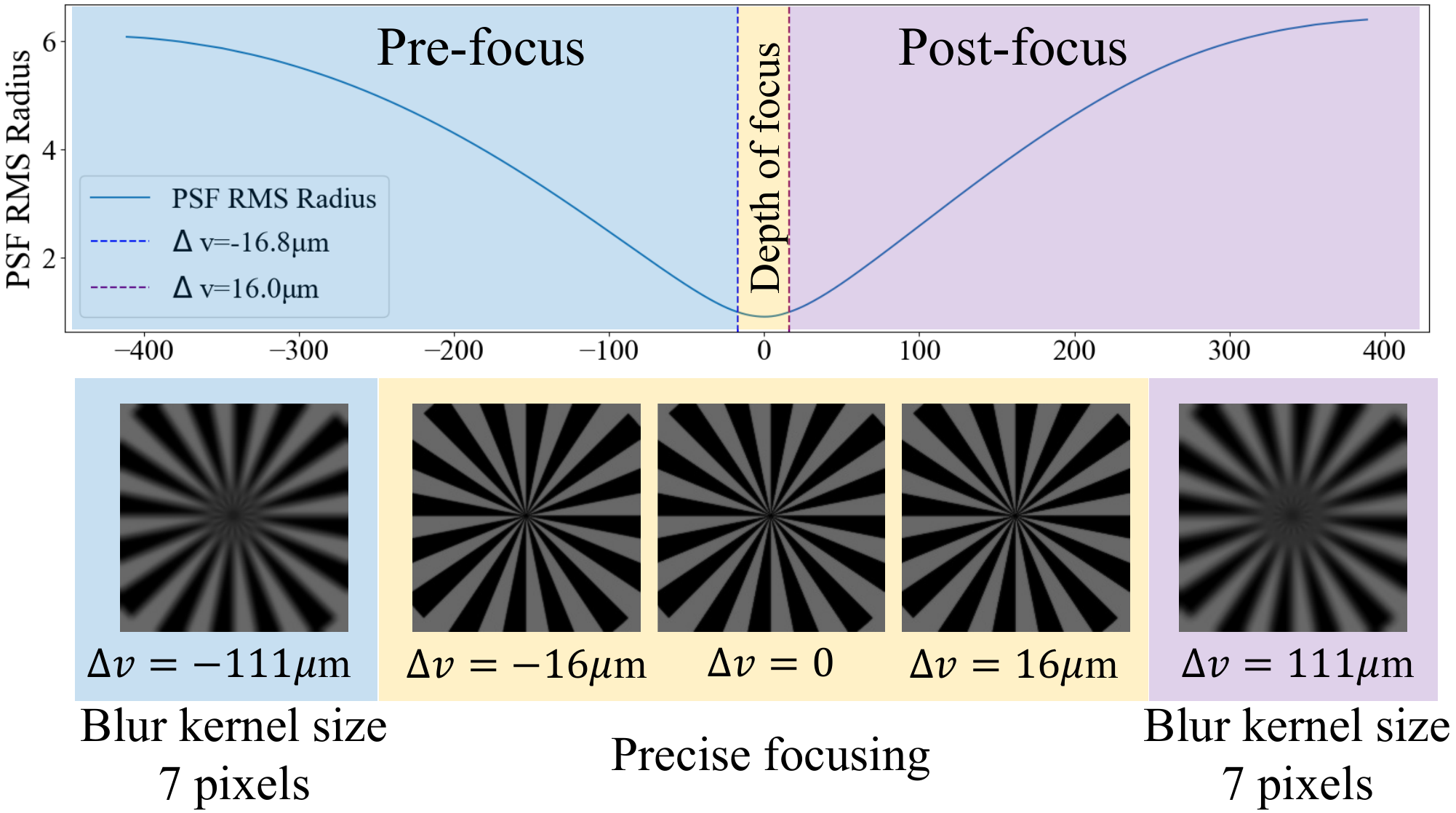}
   \caption{Top row: PSF RMS radius changes with $\Delta v$. Bottom row: the visual explanation of the depth of focus. The yellow box represents the depth of focus region, while the blue and purple boxes indicate the pre-focus and post-focus regions, respectively.}
   \label{fig:dof}
\end{figure*}

The concept of ``depth of focus'' has been referenced multiple times in the main text. Precise focusing is defined as a focusing error within one depth of focus. For example, considering the lens used for the synthetic dataset with a depth of focus of 16 $\mu$m, the blur kernel can be approximated as a Dirac function, $\delta(\boldsymbol{x})$, whenever the focusing error falls within this range. As illustrated in the bottom row of \cref{fig:dof}, the images within the yellow box are indistinguishable from the sharp original image, indicating precise focusing.

However, aiming for a smaller focus error within the depth of focus offers clear benefits: it brings the lens closer to the center depth within the focus ROI, allowing objects just in front of and behind the main focus point to stay sharp. For instance, when focusing on a face, it’s preferable for features like the nose tip and ears to be as clear as the eyes, rather than achieving sharpness for the eyes and nose while letting the ears become blurred.

\section{Event-only one-step AF details}

In the event-only one-step AF system, Event-driven Temporal-mapping Photography (EvTemMap)~\cite{bao2024temporal} plays a crucial role by providing ELP with a single grayscale reference frame for accurate Laplacian computation. As discussed in~\cref{sec:real_result}, EvTemMap images exhibit a high dynamic range, ultra-high grayscale resolution, and an extended depth of field, all of which facilitate precise Laplacian acquisition in ELP. The high dynamic range and grayscale resolution arise from temporal mapping, where each microsecond timestamp is mapped to one grayscale level, resulting in nearly 20,000 levels over a 20 ms exposure. The extended depth of field is achieved through a specialized transmittance modulation approach: in our setup, an aperture shutter opens progressively from fully closed. In this setup, brighter areas correspond to smaller apertures, which, in turn, produce a greater depth of field.

Comparing Scene 2 in \cref{fig:davis} with Scene 1 in \cref{fig:evk4}, the dynamic range of the grayscale image captured by the DAVIS346 APS sensor is notably lower than that of the EvTemMap image, which reveals detailed texture. The higher grayscale resolution in EvTemMap provides more refined Laplacian information for ELP, resulting in generally higher ELP values for the EVK4 dataset compared to the DAVIS dataset. Additionally, the extended depth of field in EvTemMap makes the defocus image appear sharper, resembling an all-in-focus image. This enhanced sharpness provides more precise texture information, thereby increasing the steepness of the ELP ``sign mutation'' at the focus position. Although the grayscale image has a large depth of field, the focus events still correspond to a shallow depth of field, ensuring accurate focus position detection. 
In a darker environment, such as Scene 2 in \cref{fig:evk4}, the defocus image from EvTemMap appears less sharp. Consequently, the ELP curve shows a slightly reduced steepness at the ``sign mutation'' point, potentially increasing focusing error. However, as in most focusing scenarios, beginning with a small defocus amount enables the event-only one-step autofocus system to function effectively, even under low-light conditions.


\section{Detailed results on the synthetic dataset}

\begin{table*}[ht]
\centering
\small
\setlength{\tabcolsep}{2pt}
\renewcommand{\arraystretch}{1}
\begin{tabular}{c|c|p{0.89cm}p{0.89cm}p{0.89cm}|p{0.89cm}p{0.89cm}p{0.89cm}|p{0.89cm}p{0.89cm}p{0.89cm}|p{0.89cm}p{0.89cm}p{0.89cm}|p{0.89cm}p{0.89cm}p{0.89cm}}

\toprule
\multirow{2}{*}{\textbf{Scene}} & \multirow{2}{*}{\textbf{FoV}} & \multicolumn{3}{c|}{\textbf{EGS}} & \multicolumn{3}{c|}{\textbf{PBF}} & \multicolumn{3}{c|}{\textbf{ELP(1FPS)}} & \multicolumn{3}{c|}{\textbf{ELP(20FPS)}} & \multicolumn{3}{c}{\textbf{ELP(50FPS)}} \\
 &  & \textbf{S} & \textbf{M} & \textbf{V} & \textbf{S} & \textbf{M} & \textbf{V} & \textbf{S} & \textbf{M} & \textbf{V} & \textbf{S} & \textbf{M} & \textbf{V} & \textbf{S} & \textbf{M} & \textbf{V} \\
\hline
 & 1 & / & / & / & 1.20 & 1.60 & 0.80 & -8.00 & -8.00 & -6.40 & -2.40 & -5.60 & -0.80 & -1.60 & -3.20 & -4.80 \\
focus- & 2 & -44.80 & / & -57.60 & 2.40 & 4.00 & 0.40 & -3.20 & -12.80 & -8.00 & -0.80 & -0.80 & -2.40 & 0.00 & -1.60 & -1.60 \\
board& 3 & -45.60 & -40.00 & -33.87 & 3.60 & 3.60 & 3.20 & -4.80 & -6.40 & -6.40 & -2.40 & -0.80 & -0.80 & -2.40 & 0.00 & 0.00 \\
 & 4 & -36.00 & -50.80 & / & 10.00 & 10.00 & 7.20 & 3.20 & 1.60 & 6.40 & -0.80 & 0.80 & 2.40 & 0.00 & 0.00 & 1.60 \\
 \hline
 & 1 & -31.20 & / & / & -2.40 & -2.80 & -3.20 & -2.40 & 2.40 & -0.80 & -1.60 & -1.60 & -6.40 & -0.80 & -2.40 & -2.40 \\
 & 2 & / & -12.80 & -6.00 & 5.60 & 0.80 & -2.40 & 0.00 & -1.60 & -3.20 & 2.40 & 0.80 & -0.80 & 3.20 & 1.60 & 1.60 \\
dove & 3 & / & / & / & 4.80 & 4.00 & 4.00 & 1.60 & 0.00 & 0.00 & 5.60 & 4.00 & 7.20 & 6.40 & 4.80 & 4.80 \\
 & 4 & / & / & -36.80 & 1.60 & 4.80 & 1.60 & 0.80 & -4.00 & -5.60 & -3.20 & -3.20 & -4.80 & -0.80 & -0.80 & -0.80 \\
 \hline
 & 1 & -38.40 & -16.80 & / & -3.20 & -3.20 & -4.00 & 0.80 & -0.80 & -4.00 & -1.60 & -4.80 & -3.20 & -2.40 & -2.40 & -4.00 \\
 & 2 & / & / & / & 6.40 & 0.80 & -4.00 & 1.60 & -3.20 & -1.60 & 0.80 & 0.80 & 0.80 & 1.60 & 1.60 & 0.00 \\
cat & 3 & -31.20 & / & -31.20 & 5.60 & 4.00 & 4.00 & 3.20 & 1.60 & 0.00 & 4.00 & 2.40 & 0.80 & 4.80 & 3.20 & 1.60 \\
 & 4 & / & / & -40.80 & 4.80 & 8.00 & 4.80 & 2.40 & 0.80 & 2.40 & 0.00 & -1.60 & -4.80 & -0.80 & -0.80 & -0.80 \\
\hline
 & 1 & -38.80 & -49.20 & 2.20 & 4.00 & 2.40 & 0.80 & 0.00 & -3.20 & -1.60 & 0.80 & -2.40 & -2.40 & 1.60 & 0.00 & -1.60 \\
 & 2 & -38.40 & -43.60 & -33.76 & 6.40 & 4.00 & -4.80 & 0.00 & -4.80 & 1.60 & 0.80 & -2.40 & 0.80 & 0.00 & -1.60 & 1.60 \\
flower & 3 & -50.40 & -50.00 & -25.92 & 4.80 & 4.00 & 2.40 & 0.00 & -1.60 & -4.80 & 2.40 & 0.80 & -0.80 & 3.20 & 1.60 & -0.80 \\
 & 4 & / & -44.40 & -38.00 & 10.40 & 10.40 & 4.00 & 4.80 & 0.00 & -1.60 & 0.80 & -0.80 & 0.80 & 1.60 & 0.00 & 0.00 \\
\hline
 & 1 & -38.80 & -15.60 & 8.64 & 4.00 & 0.80 & -0.80 & 0.00 & 3.20 & 1.60 & 0.80 & -4.00 & -4.00 & 0.00 & -1.60 & -3.20 \\
 & 2 & -36.80 & -34.40 & 29.87 & 7.20 & -2.40 & -4.00 & 0.00 & -8.00 & -9.60 & 0.80 & -2.40 & 0.80 & 1.60 & 0.00 & 4.80 \\
leaf & 3 & -40.80 & / & -32.40 & 7.20 & 4.00 & -5.60 & 0.00 & -1.60 & -6.40 & 4.00 & 0.80 & 0.80 & 6.40 & 3.20 & 4.80 \\
 & 4 & -29.60 & -29.60 & 14.20 & 11.20 & 13.60 & 2.40 & 4.80 & 0.00 & 9.60 & -0.80 & -0.80 & 7.20 & 1.60 & 0.00 & 0.00 \\
\hline
& 1 & -4.27 & -0.16 & 7.73 & 0.40 & 0.80 & 1.20 & 1.60 & -1.60 & 0.00 & -0.80 & -5.60 & -8.80 & -1.60 & -3.20 & -3.20 \\
 & 2 & -5.80 & -5.71 & 7.70 & -0.80 & -0.80 & 0.00 & -1.60 & -4.80 & -4.80 & -2.40 & -4.00 & -4.00 & -3.20 & -3.20 & -3.20 \\
chair  & 3 & -29.76 & -34.40 & -7.36 & 0.80 & 1.60 & 0.80 & -3.20 & -6.40 & -3.20 & -2.40 & -0.80 & -4.00 & -0.80 & -0.80 & 0.00 \\
 & 4 & -3.20 & -3.60 & -14.40 & 4.00 & 4.00 & 1.20 & 0.00 & -3.20 & 6.40 & -2.40 & -4.00 & -4.00 & -1.60 & -1.60 & -3.20 \\
\hline
 & 1 & / & -21.20 & 15.60 & 4.00 & 0.00 & 0.80 & 3.20 & -3.20 & -1.60 & 2.40 & -4.00 & -0.80 & 1.60 & -1.60 & -1.60 \\
 & 2 & -37.60 & -32.00 & -1.60 & 7.20 & -5.60 & -4.00 & 1.60 & -8.00 & 1.60 & 4.00 & -2.40 & 5.60 & 3.20 & 0.00 & -1.60 \\
 grass& 3 & -32.00 & -30.67 & 11.73 & 3.60 & -2.40 & -2.00 & -1.60 & -4.80 & -1.60 & 2.40 & -0.80 & -0.80 & 3.20 & 0.00 & 4.80 \\
 & 4 & 52.80 & -42.40 & / & 10.40 & 7.20 & 0.80 & 1.60 & -4.80 & 1.60 & -2.40 & -4.00 & -2.40 & 0.00 & -1.60 & 4.80 \\
\bottomrule
\end{tabular}
\caption{Detailed results of the synthetic dataset, measured in $\mu$m. S for Static, M for Moderate motion, V for Violent motion. `/' means that focus position fails to be identified.}
\label{tab:syn_detail}

\end{table*}

\Cref{tab:syn_detail} details the focusing errors of the EGS, PBF, and ELP methods for the 84 scenes of the synthe dataset. In the synthetic dataset, the ground truth for the focus position is the point with the smallest PSF RMS radius. With a focus depth of 16 $\mu$m, the focus is considered accurate if the focus error remains within 16 $\mu$m and the blur kernel radius is smaller than one pixel. 
As shown in~\cref{tab:syn_detail}, all three ELP setups produce accurate focus results, as does the PBF. In contrast, the EGS achieves accurate focus in only 17.8\% of cases, and in 28.6\% of cases, it fails to provide any valid focus results (denoted by `/').

\section{Ablation study on reconstruction quality.}
\begin{figure*}[ht]
  \centering
   \includegraphics[width=0.99\linewidth]{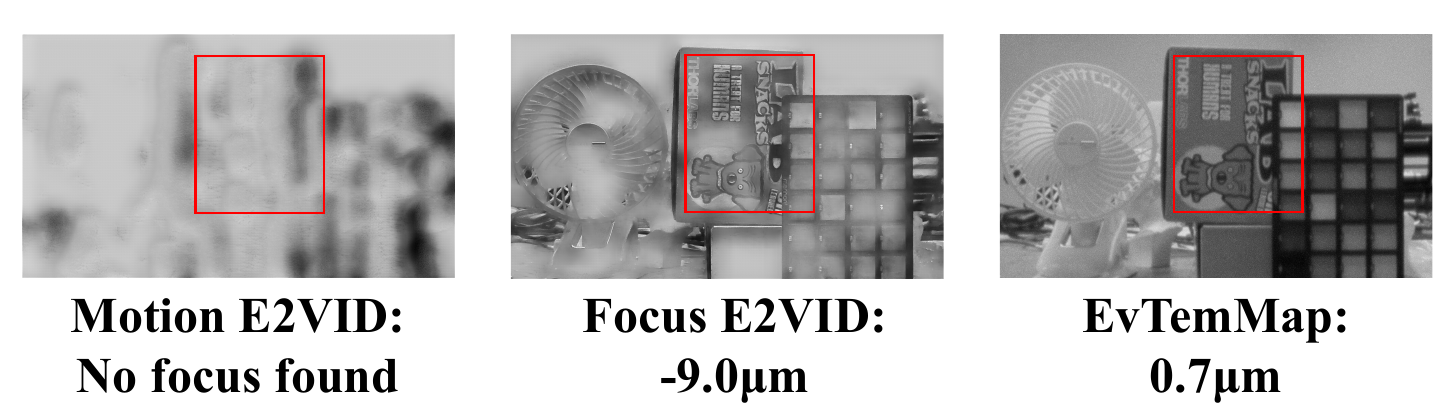}
   \caption{Ablation study on reconstruction quality.}
   \label{fig:e2vid}
\end{figure*}
The quality of image reconstruction greatly impacts ELP. We compare two approaches: Motion E2VID, which applys E2VID to the ego-motion events collected before the focus stage, and Focus E2VID, which applys E2VID to focus events of the entire stack. Frames with the lowest NIQE are selected as reference frames for ELP.
As shown in \cref{fig:e2vid}, Motion E2VID fails due to degraded texture, while Focus E2VID, though improved, introduces grayscale errors, increasing MAE on the EVK4 dataset from 0.57$\mu$m to 43.2$\mu$m. In contrast, EvTemMap not only provides higher-quality reference frames, but also offers a more streamlined workflow: Opening a closed aperture before capturing a snapshot aligns better with user habits than inducing ego-motion or pre-capturing a full focus stack.







\end{document}